\documentclass[10pt,twocolumn,letterpaper]{article}

\usepackage[pagenumbers]{cvpr} 

\definecolor{cvprblue}{rgb}{0.21,0.49,0.74}
\usepackage[pagebackref,breaklinks,colorlinks,allcolors=cvprblue]{hyperref}
\usepackage{booktabs}
\usepackage{colortbl}
\usepackage{graphicx} 
\usepackage{amsmath} 
\definecolor{cvprblue}{rgb}{0.21,0.49,0.74}
\usepackage[pagebackref,breaklinks,colorlinks,allcolors=cvprblue]{hyperref}
\usepackage{booktabs}
\usepackage{colortbl}
\usepackage{graphicx}
\usepackage{subcaption}
\usepackage{amsmath}
\usepackage[most]{tcolorbox}
\usepackage{tcolorbox}
\tcbuselibrary{breakable,skins}
\usepackage{tcolorbox}
\definecolor{boxblue}{RGB}{235, 245, 250}
\definecolor{frameblue}{RGB}{70, 130, 180}
\definecolor{boxgreen}{RGB}{240, 250, 240}
\definecolor{framegreen}{RGB}{60, 140, 60}
\definecolor{boxorange}{RGB}{255, 248, 240}
\definecolor{frameorange}{RGB}{200, 100, 50}
\definecolor{boxpurple}{RGB}{245, 240, 255}
\definecolor{framepurple}{RGB}{100, 50, 160}

\usepackage{tcolorbox}
\tcbuselibrary{breakable, skins} 
\tcbset{
    promptbox/.style={
        enhanced,
        breakable,
        colback=white,
        colframe=black,
        fonttitle=\bfseries,
        colbacktitle=black!10!white,
        coltitle=black,
        attach boxed title to top left={xshift=5mm,yshift=-2mm},
        boxrule=1pt,
        left=5mm, right=5mm, top=5mm, bottom=5mm,
        arc=2mm
    }
}
\def\paperID{284} 
\def\confName{CVPR}
\def\confYear{2026}

\title{SketchVL: Policy Optimization via Fine-Grained Credit Assignment \\ for Chart Understanding and More}

\author{
    Muye Huang\textsuperscript{\rm 1,2,3}, 
    Lingling Zhang\textsuperscript{\rm 1,2}\thanks{Corresponding author.}, 
    Yifei Li\textsuperscript{\rm 1,2,3}, 
    Yaqiang Wu\textsuperscript{\rm 4}, 
    Jun Liu\textsuperscript{\rm 1,2}
    \vspace{0.5em} \\ 
    \textsuperscript{\rm 1}School of Computer Science and Technology, Xi’an Jiaotong University, China \\
    \textsuperscript{\rm 2}MOE KLINNS Lab, Xi’an Jiaotong University, China \\
    \textsuperscript{\rm 3}Zhongguancun Academy, Beijing, China \quad
    \textsuperscript{\rm 4}Lenovo Research \\
    {\tt\small \{huangmuye, liyifei619584902\}@stu.xjtu.edu.cn} \\
    {\tt\small \{zhanglling, liukeen\}@xjtu.edu.cn, wuyqe@lenovo.com}
}

\begin{document}
\maketitle
\begin{abstract}
Charts are high-density visual carriers of complex data and medium for information extraction and analysis. Due to the need for precise and complex visual reasoning, automated chart understanding poses a significant challenge to existing Multimodal Large Language Models (MLLMs). Many MLLMs trained with reinforcement learning (RL) face the challenge of credit assignment. Their advantage estimation, typically performed at the trajectory level, cannot distinguish between correct and incorrect reasoning steps within a single generated response. To address this limitation, we introduce SketchVL, a novel MLLM that optimized with FinePO, a new RL algorithm designed for fine-grained credit assignment within each trajectory. SketchVL's methodology involves drawing its intermediate reasoning steps as markers on the image and feeding the annotated image back to itself, creating a robust, multi-step reasoning process. During training, the FinePO algorithm leverages a Fine-grained Process Reward Model (FinePRM) to score each drawing action within a trajectory, thereby precisely assigning credit for each step. This mechanism allows FinePO to more strongly reward correct tokens when a trajectory is globally successful, and more heavily penalize incorrect tokens when the trajectory is globally suboptimal, thus achieving fine-grained reinforcement signals. Experiments show that SketchVL learns to align its step-level behavior with the FinePRM, achieving an average performance gain of 7.23\% over its base model across chart datasets, natural image datasets, and mathematics, providing a promising new direction for training powerful reasoning models. 
\end{abstract}    
\section{Introduction}
\label{sec:intro}
Charts, as a primary method of data visualization, are capable of presenting data with accuracy and intuition. They are widely found in technical documents, business reports, and scientific papers. The automated understanding of charts is instrumental in advancing the automated analysis of documents, and its significant research value has progressively drawn widespread attention from researchers. Recent advanced Multimodal Large Language Models, including Qwen2.5VL ~\cite{bai2025qwen25vltechnicalreport}, Gemma3~\cite{gemmateam2025gemma3technicalreport}, GPT-5~\cite{openai2024gpt4technicalreport}, and Gemini-2.5~\cite{comanici2025gemini25pushingfrontier}, have shown numerous improvements in chart understanding. Concurrently, RL in the multimodal domain, demonstrated by approaches like Vision-R1~\cite{DBLP:journals/corr/abs-2503-06749} and VLM-R1~\cite{DBLP:journals/corr/abs-2504-07615}, has been shown to effectively enhance models' multimodal understanding capabilities.


\begin{figure*}
    \centering
    \includegraphics[width=1\linewidth]{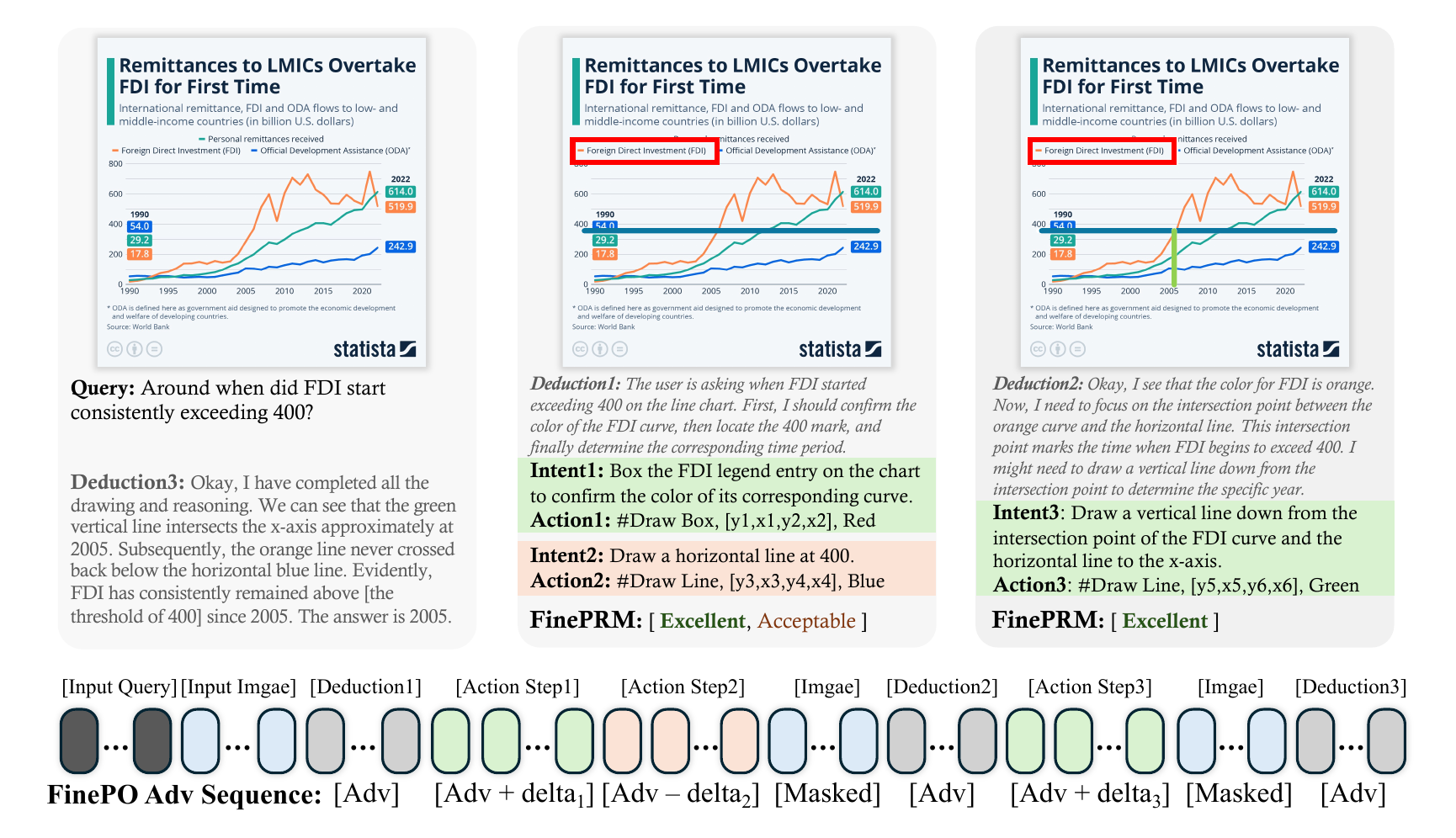}
    \caption{Illustration of our SketchVL FinePO process. SketchVL first decomposes a complex query into a trajectory of visual actions, which are scored by a FinePRM. FinePO then achieves credit assignment by redistributing credit based on the FinePRM scores. Here, $Delta_{1-3}$ represent the process rewards for Actions 1-3, calculated from their FinePRM scores, and the preceding sign indicates whether an action is considered relative advantaged or disadvantaged within the trajectory. The detailed calculation is described in Section~\ref{sec:formatting}.}
    \label{main_pic}
\end{figure*}


Building on these advances, chart understanding itself exposes a distinctive, stepwise structure: models must first localize legends and axes, then read values, align categories, compare trends, and finally synthesize conclusions. Any minor slip, such as an imprecise crop, a misread tick, or a mismatched legend, can derail the entire chain of reasoning. This step dependency makes chart analysis particularly sensitive to how reinforcement signals are assigned.

However, most prevailing RL practices for MLLMs still deliver coarse, outcome-only feedback. Methods such as GRPO~\cite{deepseekai2025deepseekr1incentivizingreasoningcapability} compute a single advantage from the final result and broadcast it uniformly to all tokens in the trajectory. When applied to charts’ multi-step pipelines, such uniform credit can inadvertently punish otherwise sound intermediate logic in an incorrect answer, or reward flawed steps simply because the final prediction happens to be right—injecting noise and limiting the benefits of RL.
These characteristics call for fine-grained credit assignment: assessing and reinforcing each intermediate step along the reasoning chain. By rewarding correct sub-decisions and precisely penalizing missteps, the learning signal becomes sharper and lower-noise, better matching the compositional nature of chart analysis.

Motivated by these observations, we introduce SketchVL, a novel multimodal interactive reasoning model. SketchVL operates within the \textit{Reasoning on Image} (RoI) paradigm, exemplified by works like ChartSketcher~\cite{DBLP:journals/corr/abs-2505-19076} and DeepEyes~\cite{DBLP:journals/corr/abs-2505-14362}. This paradigm requires the model to externalize its intermediate reasoning steps into a visible trajectory of marking actions on the image. This explicit, step-by-step decomposition of the reasoning process provides the necessary structure for step-level credit assignment. We optimize SketchVL using our FinePO algorithm, as shown in Figure~\ref{main_pic}, which is specifically designed to operate on these action trajectories. FinePO first computes a relative advantage score for each complete trajectory by making comparisons within a group, an approach inspired by GRPO. It then introduces its central mechanism of secondary credit redistribution. This process leverages a specially designed Fine-grained Process Reward Model (FinePRM) to evaluate the quality of each individual action within the trajectory. Based on the FinePRM scores, FinePO ultimately redistributes the trajectory's overall relative advantage among its constituent actions according to their contributions.


To implement SketchVL, we followed the common `ColdStart-RL' training paradigm. For the cold start phase, we reused parts of the ChartSketcher data pipeline to construct a dataset of 50K samples. For the subsequent reinforcement learning phase, we collected 9K mixed-domain data points. Concurrently, to support the FinePO algorithm, we built a sophisticated cross-modal distillation pipeline to generate training data for the FinePRM. This pipeline first decomposes dense visual information from charts into structured textual annotations. A LLM is then used to distill these annotations into a large-scale set of `intent-action' multimodal training pairs, complete with both positive and negative samples. In total, we collected 473K data points through this method, specifically for training the FinePRM's capability to assess process quality.


Our main contributions are threefold:
\begin{itemize}
    \item We introduce SketchVL, a powerful MLLM based on the \textit{Reasoning on Image} paradigm. We also contribute two large-scale datasets: a 473K multimodal dataset for training FinePRM, and a 50K dataset for RoI cold start.
    \item We propose FinePO, a novel reinforcement learning algorithm that addresses the coarse reward problem in conventional RL through fine-grained, step-level credit assignment. By leveraging a dedicated process reward model FinePRM, FinePO provides a more precise and stable learning signal for policy optimization.
    \item We conduct extensive experiments across multiple benchmarks, demonstrating the effectiveness of FinePO. Our ablation studies further underscore the important role of credit assignment in enhancing RL performance.
\end{itemize}

\section{Related Work}
\textbf{Chart Understanding.}
Research in Chart Understanding focuses on interpreting the visual context of charts for tasks such as question answering and summarization. The seminal work of FigureQA~\cite{DBLP:conf/iclr/KahouMAKTB18} presented a pipeline for chart comprehension, addressing binary classification for related questions. Subsequent research advanced these capabilities through the use of multi-component systems~\cite{DBLP:conf/emnlp/MasryKLHJ23, DBLP:conf/icml/LeeJTH0EKSCT23, DBLP:conf/acl/0001PKPLJACE23, DBLP:conf/eccv/LevyBL22,DBLP:journals/corr/abs-2406-18521}. DePlot~\cite{zhou2023chartt5}, for example, utilized several components alongside the mathematical prowess of LLMs to improve results on the PlotQA benchmark~\cite{DBLP:conf/wacv/MethaniGKK20}.

The advent of MLLMs has shifted the paradigm, leading to MLLM-centric approaches becoming the mainstream~\cite{DBLP:journals/corr/abs-2403-09028, masry2024chartgemmavisualinstructiontuningchart, DBLP:journals/corr/abs-2404-09987, DBLP:conf/aaai/HuangLZW0ZL25, DBLP:journals/corr/abs-2402-12185, DBLP:journals/corr/abs-2309-11268,zhang2025pixelcraftmultiagenthighfidelityvisual}. By strategically constructing training data and fine-tuning LLaVA~\cite{li2024llava}, ChartLlama~\cite{DBLP:journals/corr/abs-2311-16483} produced a capable chart-expert model. Capitalizing on the inherent language capabilities of MLLMs, recent work has adopted multi-task training to enhance chart comprehension. ChartAssistant~\cite{DBLP:journals/corr/abs-2401-02384}, for instance, employed unified multi-task training for better overall performance. The Program-of-Thoughts technique has been applied by TinyChart~\cite{DBLP:journals/corr/abs-2404-16635} to boost numerical reasoning, while ChartMoE~\cite{xu2025chartmoemixturediverselyaligned} adopted a Mixture of Experts architecture to handle diverse chart formats.

\textbf{Reasoning and RL.}
Models like OpenAI-o1~\cite{openai2024openaio1card} and Deepseek-R1~\cite{deepseekai2025deepseekr1incentivizingreasoningcapability} have showcased the potent reasoning abilities of LLMs~\cite{grattafiori2024llama3herdmodels, qwen2, qwen2025qwen25technicalreport, xu2024interactiveevolutionneuralsymbolicselftraining, qwq32b, xu2025geniusgeneralizablepurelyunsupervised}, which are frequently improved via RL. Nevertheless, reasoning within MLLMs remains a subject of active research. Many contemporary methods center on employing Chain-of-Thought (CoT) techniques~\cite{DBLP:conf/nips/Wei0SBIXCLZ22, DBLP:conf/nips/YaoYZS00N23} to train MLLMs for generating step-by-step inference sequences. Such approaches~\cite{DBLP:journals/corr/abs-2411-14432, DBLP:conf/cvpr/LiYLMZYSLB24, DBLP:journals/corr/abs-2403-12966, cheng2025caparenabenchmarkinganalyzingdetailed} primarily focus on CoT within the textual modality, depending heavily on the MLLM's language backbone.
To better integrate visual cues, VisualCoT~\cite{DBLP:conf/nips/ShaoQ0SZW0024} proposed cropping critical regions to guide the model's focus. A more recent direction, \textit{Reasoning on Image}, involves models externalizing their reasoning by generating visual markers on the image itself, creating an interactive feedback loop~\cite{DBLP:journals/corr/abs-2501-07542,DBLP:journals/corr/abs-2505-08617,DBLP:journals/corr/abs-2505-11409,DBLP:journals/corr/abs-2505-22596,DBLP:journals/corr/abs-2505-15966,DBLP:journals/corr/abs-2501-05452}. More recently, attention has also turned to the credit assignment problem in reinforcement learning as a means to refine the policy optimization process for reasoning tasks~\cite{guo2025segmentpolicyoptimizationeffective}.
\section{Method}
\label{sec:formatting}

We introduce SketchVL, a MLLM that employs an iterative reasoning process. During its reasoning, the model explicitly expresses its thought process by rendering visual markers onto the chart. This annotated chart is then fed back to the model itself to guide the subsequent decision, thereby forming a visible reasoning trajectory. We optimize SketchVL using the novel FinePO, a reinforcement learning method we designed to achieve step-level, fine-grained credit assignment. FinePO leverages our designed Fine-grained Process Reward Model (FinePRM) to evaluate the quality of each marking action along the reasoning path, which in turn enables precise credit assignment. In the following sections, we will detail the training methodology for SketchVL, the workflow of FinePO, and the construction details of the FinePRM.

\subsection{SketchVL}
The training of SketchVL follows a two-stage methodology: it first learns to generate interactive reasoning sequences in a Cold Start phase, followed by optimization in a FinePO RL phase.
During the cold start phase, the model is supervised to acquire foundational localization capabilities and robust reasoning patterns for reasoning-on-image tasks. The subsequent RL phase then employs our FinePO algorithm to unlock the model's more complex reasoning abilities. The training data for the Cold Start phase is sourced from two main streams: the ground-truth steps generated by the `\textit{Trajectorie-based Simulation}' pipeline (as will be described in FinePRM data collection), and the ChartSketcher data pipeline. A detailed breakdown of the data composition is provided in the Experiments section.

\subsection{FinePO}
FinePO is a reinforcement learning method that enables step-level, fine-grained credit assignment. To achieve this, it operates on a reasoning trajectory that is decomposed into a sequence of discrete, visual marking actions. Each step in this trajectory consists of a textual `intent', which describes the reasoning goal (e.g., ``mark the maximum value"), and a corresponding `action', which executes this intent by rendering a visual marker on the image. This mechanism of explicitly externalizing the thought process into visible steps is fundamental to our approach.

The FinePO algorithm can be conceptualized in two main phases. First, we compute a coarse, cross-trajectory advantage by comparing the overall quality of different generated responses, following the approach of GRPO. Second, we perform an intra-trajectory credit assignment, where this coarse advantage signal is meticulously redistributed among the fine-grained steps that constituted the response.

\subsubsection{Cross-trajectory advantage compute}
For a given prompt, we generate a set of $k$ candidate responses, $\{y_1, y_2, \dots, y_k\}$, which collectively form a ``trajectory". Each response $y_i$ is assigned a terminal reward $R(y_i)$ based on an evaluation of its overall correctness. The advantage $A(y_i)$ for each response is then calculated by subtracting the mean reward of the comparison group from the individual response's reward:
\begin{equation}
A(y_i) = R(y_i) - \frac{1}{k} \sum_{j=1}^{k} R(y_j)
\label{eq:advantage}
\end{equation}

This advantage value, $A(y_i)$, indicates whether a particular response $y_i$ is better ($A(y_i) > 0$) or worse ($A(y_i) < 0$) than the average performance within its trajectory group. This scalar advantage serves as the coarse, high-level signal that will subsequently be distributed among the fine-grained steps in the next phase.

\subsubsection{Intra-trajectory credit assignment}
FinePO introduces an intra-trajectory credit assignment mechanism. This mechanism utilizes a FinePRM to assess the quality of each individual step, allowing rewards and penalties to be applied with greater precision to the steps themselves, rather than being determined solely by the trajectory's final outcome.
To achieve this, we first define our FinePRM, denoted as $\mathcal{P}$, as a function that provides a scalar score for each reasoning step based on its intent and the visual change it produces:
\begin{equation}
p_j = \mathcal{P}(\text{intent}_j, \text{action}_j, \text{img}_{j-1}, \text{img}_j)
\end{equation}
where $p_j$ is the process score for step $s_j$, and $\text{img}_{j-1}$ and $\text{img}_j$ are the images before and after the action is rendered.

However, the policy may develop a preference for action types that are easier to score highly, while avoiding actions that are important but potentially harder to execute perfectly. This bias arises from applying a unified scoring standard to action types that have varying levels of intrinsic difficulty.

To counteract this bias, we introduce a KL divergence constraint to penalize deviations between the model's generated action distribution and a prior distribution observed in the training set. Specifically, for a step $s_j$ with action type $a_j$, we first compute a clipped KL penalty offset:
\begin{equation}
    \mathcal{O}_{\text{clipped}}(a_j) = \text{clip}\left(-\lambda_{\text{KL}} \log \left( \frac{P_k(a_j) + \epsilon}{Q(a_j) + \epsilon} \right), -\gamma, \gamma\right)
\end{equation}
where $Q(a)$ is the pre-computed prior action distribution from the training set, and $P_k(a)$ is the current policy's action distribution computed over a sliding window of the last $k$ batches. $\lambda_{\text{KL}}$ is the KL penalty coefficient, $\gamma$ is the clipping threshold, and $\epsilon$ ensures numerical stability.

This offset is then used to adjust the original process score, yielding a regularized score $p'_j$:
\begin{equation}
    p'_j = p_j + \mathcal{O}_{\text{clipped}}(a_j)
\end{equation}

To perform credit assignment, we next convert these regularized, absolute scores $p'_j$ into values relative to the intra-response average. This is achieved by calculating a weighted mean. Specifically, for an entire response $y_i$, we weight each step's score $p'_j$ by its corresponding token length $L_j$:
\begin{equation}
\bar{p} = \frac{\sum_{j=1}^{N} L_j \cdot p'_j}{\sum_{j=1}^{N} L_j}
\end{equation}
The deviation for each step, $\Delta_j = p'_j - \bar{p}$, then represents whether a step is of higher or lower quality than the intra-response average.

\begin{figure*}
    \centering
    \includegraphics[width=1\linewidth]{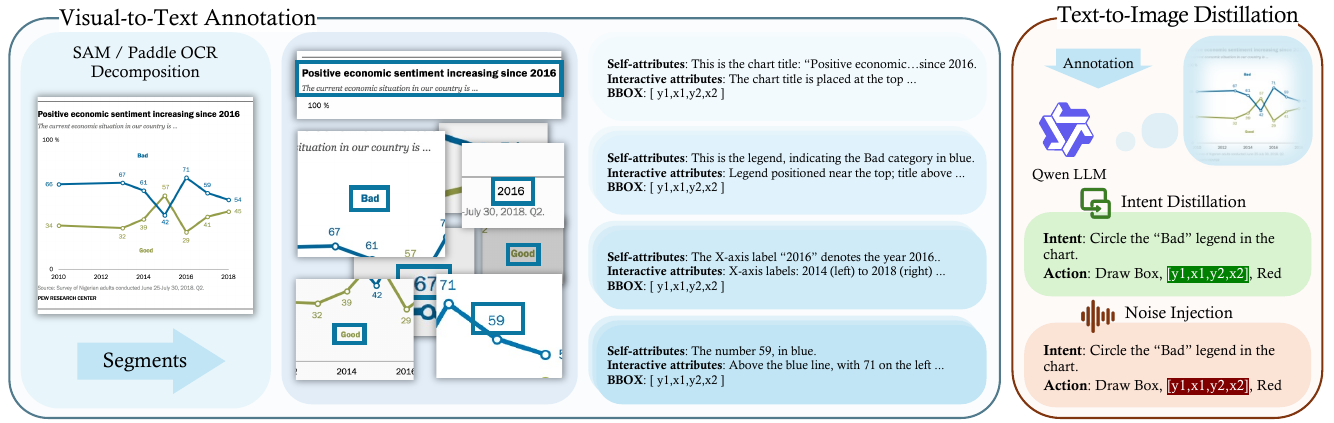}
    \caption{The data pipeline for training our FinePRM. First, we convert charts into structured textual annotations via decomposition and description (left). Second, we use LLMs to distill these annotations into a training dataset of `intent-action' pairs, including both positive samples and negative samples generated via noise injection (right).}
    \label{fig:placeholder}
\end{figure*}

Our goal is not to create new rewards, but to redistribute the existing coarse advantage $A(y_i)$ more accurately among its constituent steps. This ensures that the fine-grained learning signal remains grounded in the empirically observed overall performance of the response. The weighted sum of the adjustments across all steps is designed to be zero, thus conserving the total advantage. We realize this redistribution with the following formula:
\begin{equation}
    A'(s_j) = A(y_i) + \alpha \cdot k \cdot \Delta_j
    \label{eq:redistribution}
\end{equation}
where $A(y_i)$ is the coarse advantage computed in Equation~\ref{eq:advantage}, $\alpha$ is a hyperparameter that controls the intensity of the credit adjustment, and $k$ is a dynamic scaling factor that makes the adjustment's magnitude proportional to $|A(y_i)|$, defined as:
\begin{equation}
    k = \frac{|A(y_i)|}{\max_{j \in \{1..N\}}(0, \Delta_j) + \epsilon}
\end{equation}
Here, $\epsilon$ is a small constant for numerical stability.

Finally, to guarantee that steps from a globally superior response ($A(y_i)>0$) do not receive negative advantages, and vice versa, we apply a clipping function to produce the final step-level advantage $A(s_j)$:
\begin{equation}
    A(s_j) = \begin{cases} 
          \text{clip}(A'(s_j), 0, \beta \cdot A(y_i)) & \text{if } A(y_i) > 0 \\
          \text{clip}(A'(s_j), \beta \cdot A(y_i), 0) & \text{if } A(y_i) \leq 0
    \end{cases}
\end{equation}
where $\beta$ is a hyperparameter that defines the bounds of the clipping range. This final step ensures a stable signal for each individual action in the policy optimization process.

\subsection{FinePRM}
The FinePRM is a process reward model that provides the signal for FinePO's fine-grained credit assignment. In the following sections, we detail the architecture of the FinePRM and the methodology for its data collection.

\subsubsection{Architecture of FinePRM}

Similar to prior work VisualPRM~\cite{DBLP:journals/corr/abs-2503-10291}, we employ a MLLM as the backbone of our FinePRM. As defined previously, the FinePRM takes as input the textual `intent' and `action', alongside the pair of images before and after the action, $\text{img}_{j-1}$ and $\text{img}_j$.

These inputs are structured using a specific prompt template that frames the task as an evaluative judgment. The MLLM is presented with both images and is guided by a textual query to act as an assessor. The query instructs the model to carefully compare the visual modification between the two images and evaluate whether this change is a precise and correct execution of the given `intent' and `action'. A conceptual representation of the query is as follows:
\begin{quote}
\scriptsize
\texttt{Image Before: [Image 1]}\\
\texttt{Image After: [Image 2]}\\
\texttt{The following modification was ... Critically evaluate if the modification in Image 2 is a precise and correct realization of this intent. Classify the quality into one of four levels: [Excellent, Acceptable, Poor, Unacceptable].}
\end{quote}
We formulate this as a four-way classification task. The FinePRM is trained to output one of the four discrete labels: \textit{Excellent}, \textit{Acceptable}, \textit{Poor}, or \textit{Unacceptable}. These categorical judgments are subsequently mapped to scalar values $[4.0, 3.0, 2.0, 1.0]$ to serve as the process score $p'_j$ required for the intra-trajectory credit assignment phase.

\subsubsection{FinePRM Training Data Collection}
Constructing the training data for FinePRM requires a large-scale dataset of `intent-action' pairs. However, even advanced MLLMs, such as Gemini 2.5 Pro, still struggle with complex visual grounding. Consequently, conventional distillation approaches are both cost-prohibitive and incapable of generating `intent-action' pairs with the required precision.
To address this issue, we employ a sophisticated cross-modal distillation method that transforms raw images into precise `intent-action' annotations. Our approach is executed in two main stages. First, we translate the visual modality of an image into a dense, intermediate representation composed of \textbf{\textit{[text description \& target location]}} annotation pairs. Subsequently, these structured annotations are processed by LLMs to distill the final multimodal training data for FinePRM.

\paragraph{1. Visual-to-Text Annotation}
MLLMs struggle to simultaneously localize and identify a large number of objects in dense images. To address this, we require precise grounding information for each object. Inspired by \textit{Set-of-Mark}~\cite{DBLP:journals/corr/abs-2310-11441}, we leverage the SAM~\cite{kirillov2023segment} as an auxiliary segmentation tool to provide this information. We first employ SAM to partition an image into numerous object-centric patches.
These patches are then fed into a leading open-source MLLM for labeling. To provide local context, we expand each patch to include an additional 20\% surrounding area from the original image.
We then center the target object within this expanded view and highlight it with a red bounding box before presenting it to the MLLM for annotation.

For each processed patch, we prompt the MLLM to generate two types of attributes for the highlighted object:

\begin{itemize}
    \item \textbf{Self-attributes:} The object's intrinsic name or properties, such as ``lid of a starfruit can" or ``a purple polyline".
    \item \textbf{Interactive attributes:} The object's relationship with its surroundings, such as ``a jar of peanut butter is to its left" or ``intersects with the green polyline".
\end{itemize}

As shown in Figure~\ref{fig:placeholder}, this approach significantly reduces the model's cognitive load, leveraging the MLLM's strength in single-object recognition to effectively convert the visual content into a rich set of textual annotations. It is worth noting that we employ PaddleOCR for text recognition, as SAM is not adept at this task.

\paragraph{2. Text-to-Image Distillation}
In this stage, we use the structured annotations generated previously to prompt LLM, distilling them into a comprehensive dataset of simulated `intent-action' pairs. For each generated pair, we employ a rendering engine to apply the `action' to the original image, thereby producing the $\text{img}_{j-1}$ (before) and $\text{img}_j$ (after) states required for FinePRM's input. To ensure the diversity of the training data, we synthesize these `intent-action' pairs through two distinct pipelines:

\begin{itemize}
    \item \textbf{Direct Generation:} The LLM is prompted to directly generate a straightforward `intent-action` pair from the annotations. This typically results in simple, single-step tasks, such as an intent to ``Mark the man with the white beard" or ``Highlight the 50\% value label on the x-axis".
    
    \item \textbf{Trajectory-based Simulation:} This pipeline is more complex and designed to mimic the reasoning process of FinePO. We first use the annotations to create a question-answering pair. A powerful LLM is then tasked to answer the question, simulating the generation of `trajectories` as seen in FinePO. This process yields multi-step, ground-truth (GT) reasoning trajectories, and we harvest these individual steps for subsequent training.
\end{itemize}

Finally, we inject noise to the `action' component of the collected GT steps. This process systematically generates samples exhibiting a mismatch between the stated `intent' and the executed `action'. The final training data for FinePRM is compiled with a specific 2:4:3:1 ratio for the \textit{Excellent}, \textit{Acceptable}, \textit{Poor}, and \textit{Unacceptable} labels, respectively. This non-uniform distribution is deliberately chosen to compel the model to focus on the critical and often subtle decision boundary between the `Acceptable' and `Poor' categories.

\section{Experiments}
\label{sec:Experiments}


\subsection{Settings}
\label{subsec:settings}

\subsubsection{Data Construction}

\textbf{Cold Start Data.}
For the cold-start phase, we selected images from the EvoChart, GQA, and ChartQA-Train datasets as metadata. For the EvoChart dataset, which contains synthesis source code, we reused the ChartSketcher data pipeline. For GQA and ChartQA-Train, we utilized the Visual-to-Text annotations generated during our FinePRM construction process. It is noteworthy that all QA pairs were synthetically generated; we did not use any QA pairs from the original training sets of these datasets. The final SFT data, totaling 50K samples, was distilled using the Qwen3-235B-A22B-Instruct-2507~\cite{yang2025qwen3technicalreport}.

\textbf{FinePRM Training Data.}
For the construction of the FinePRM, we used images from EvoChart~\cite{DBLP:conf/aaai/HuangLZW0ZL25}, GQA~\cite{DBLP:conf/cvpr/HudsonM19}, ChartQA-Train~\cite{DBLP:conf/acl/MasryLTJH22}, and OpenImages~\cite{DBLP:journals/corr/abs-1811-00982} as data sources. The data pipeline employed SAM~\cite{kirillov2023segment} as a segmentation tool, PaddleOCR V5~\cite{cui2025paddleocr30technicalreport} as an OCR tool, Qwen2.5VL-72B-Instruct~\cite{bai2025qwen25vltechnicalreport} for Visual-to-Text Annotation, and Qwen-NEXT-80B-A3B for Text-to-Image Distillation. We defined five action types: \texttt{Line}, \texttt{Point}, \texttt{Rectangle}, \texttt{Circle}, and \texttt{Text}. The \texttt{Text} action is excluded from credit assignment, and the data for the remaining four actions is balanced with a 1:1:1:1 ratio. The distribution for the \textit{Excellent}, \textit{Acceptable}, \textit{Poor}, and \textit{Unacceptable} labels across all data was set to a 2:4:3:1 ratio. In total, we collected 473K SFT samples to train the FinePRM's capability to assess process quality.

\textbf{RL Data.}
In the FinePO RL phase, we aggregated 9k prompts from five distinct data sources for online trajectory generation. The composition is as follows: 2k from ChartQA-Train-Augmented, 2k from ChartQA-Train-Human, 1k from Vision-R1-RL, 2k from ChartBench-Train~\cite{DBLP:journals/corr/abs-2312-15915} , and 2k from VisualCoT-Train.

\subsubsection{Evaluation}
\textbf{Benchmarks:}
We conducted extensive experiments on both Chart Expert and General-Purpose datasets.
\begin{itemize}
    \item \textbf{Chart Expert Datasets:} EvoChart-QA, ChartQA, ChartQA-Pro~\cite{masry2025chartqaprodiversechallengingbenchmark} , ChartBench, and PlotQA~\cite{DBLP:conf/wacv/MethaniGKK20}.
    \item \textbf{General-Purpose Datasets:} MMStar and MathVista.
\end{itemize}

\textbf{Evaluation Protocol:}
We employ DeepSeek-R1-Distill-Qwen-14B as an evaluation discriminator. The evaluation rules for each dataset are provided as a prompt, and the correctness of a response is determined by a 9-vote majority.

\subsubsection{Training Settings}
We trained two versions of SketchVL, based on Qwen2.5VL-7B-Instruct and Qwen2.5VL-3B-Instruct, respectively. The FinePRM was trained using Qwen2.5VL-7B-Instruct as its backbone. All experiments were conducted using the ms-swift v3.9.0.dev0~\cite{DBLP:conf/aaai/ZhaoHHWMZJWAWZC25} framework. The key hyperparameters for the FinePO reinforcement learning phase are detailed in the Table~\ref{tab:hyperparams}. FinePRM training for 4 epochs, cold start for 2 epochs and FinePO training for 1 epochs. All models were trained on 16 x NVIDIA A800 (40G) GPUs.

\begin{table}[h]
    \centering 
    \caption{Key hyperparameters for the FinePO RL phase.}
    \small
    \label{tab:hyperparams}
    \begin{tabular}{lcc}
        \toprule
        \textbf{Symbol} & \textbf{Meaning} & \textbf{Value} \\
        \midrule
        $k$ & Number of generations per prompt & 24 \\
        $\lambda_{\text{KL}}$ & KL penalty coefficient & 0.1 \\
        $\gamma$ & Clipping threshold for KL offset & 0.5 \\
        $\alpha$ & Credit adjustment intensity & 0.2 \\
        $\beta$ & Clipping range factor for advantage & 2.0 \\
        \midrule
        - & Learning Rate & 1e-6 \\
        - & Temperature & 1.0 \\
        - & GRPO KL beta & 0.01 \\
        \bottomrule
    \end{tabular}

\end{table}

\begin{table*}[t]
\centering
\caption{Performance comparison of SketchVL with other leading models and results of our ablation study across a selection of chart-expert and general-purpose benchmarks. *\textit{Random} means that, during training, random rewards are used in place of the FinePRM.}
\small

\label{tab:main_results}
\begin{tabular*}{\textwidth}{@{\extracolsep{\fill}}lccccccc}
\toprule
Model  & EvoChart-QA & ChartQA & ChartQA-Pro & ChartBench & PlotQA & MathVista & MMStar\\
\midrule
\rowcolor[gray]{0.9}
\multicolumn{8}{l}{\textit{Performance Comparison}} \\
\midrule
VLM-R1 & 40.32 & 72.98 & 39.58 & 39.58 & 54.40 & 55.10 & 48.27 \\
ChartSketcher-2B & 26.72 & 68.24 & - & 30.10 & 41.12 & - & - \\
Qwen2.5VL-7B & \underline{54.80} & \underline{82.00} & \underline{52.40} & \underline{64.78} & \textbf{63.44} & \underline{61.40} & \underline{56.67} \\
Qwen2.5VL-3B & 39.36 & 61.88 & 37.73 & 56.20 & 42.88 & 49.50 & 43.53 \\
\midrule
SketchVL-7B (Ours) & \textbf{58.64} & \textbf{83.96} & \textbf{52.62} & \textbf{65.11} & \underline{55.84} & \textbf{63.50} & \textbf{57.13} \\
SketchVL-3B (Ours) & 47.28 & 77.20 & 44.15 & 59.96 & 48.32 & 53.80 & 51.00 \\
\midrule
\rowcolor[gray]{0.9}
\multicolumn{8}{l}{\textit{Ablation Study (on SketchVL-3B)}} \\
\midrule
SketchVL-3B (Full Model) & 47.28 & \underline{77.20} & \textbf{44.15} & \textbf{59.96} & \textbf{48.32} & \underline{53.90} & \textbf{52.13} \\
 w/o FinePO (only cold start) & 42.08 & 71.80 & 38.45 & 54.57 & 35.44 & 47.60 & 49.40 \\
 w/o FinePO (naive GRPO) & 45.60 & 75.12 & 43.69 & \underline{59.26} & 44.72 & 53.80 & 49.87 \\
 w/o FinePRM (random* scores) & \underline{48.08} & 76.76 & \underline{43.94} & 58.98 & 46.40 & 52.30 & 50.87 \\
 w/o KL Action Regularization & \textbf{48.56} & \textbf{77.80} & 43.33 & 58.24 & \underline{48.16} & \textbf{54.60} & \underline{51.00} \\
 w/o Sketch (zero GRPO) & 30.48 & 57.56 & 25.92 & 53.06 & 31.12 & 45.40 & 43.67 \\
 w/o RL (use SFT) & 26.48 & 54.72 & 20.02 & 50.24 & 27.44 & 43.90 & 38.33 \\
\bottomrule
\end{tabular*}
\end{table*}

\subsection{Performance Comparison}
\label{subsec:performance}
As shown in Table~\ref{tab:main_results}, SketchVL demonstrates promising performance across multiple benchmarks.

First, our models significantly outperform their base models. SketchVL-7B surpasses Qwen2.5VL-7B on almost all chart-expert datasets, validating the effectiveness of the FinePO training method. This performance lift is even more pronounced on the 3B-scale model; SketchVL-3B achieves substantial gains over its base, Qwen2.5VL-3B, outperforming it by 15.32 and 3.76 percentage points on tasks like ChartQA and ChartBench, respectively. It is noteworthy that the magnitude of this improvement is more pronounced for the 3B model than for the 7B. We hypothesize this is because the smaller model has less capacity to ``hack" the large-scale FinePRM, compelling it to more faithfully learn the intended reasoning process from signals.

Second, SketchVL also performs exceptionally well when compared against reasoning models of a similar scale. For instance, our SketchVL-3B substantially outperforms VLM-R1 across most listed benchmarks. When compared to ChartSketcher, another leading interactive reasoning model, our model significantly surpasses its 2B version. 

Finally, it is noteworthy that SketchVL maintains strong general-purpose multimodal capabilities while enhancing its chart understanding skills. On MathVista and MMStar, two non-chart-centric datasets, SketchVL-7B achieves results that are competitive with or superior to leading models. This indicates that FinePO not only boosts performance on specialized tasks but also successfully preserves the model's generalization abilities.

\begin{figure*}[t]
    \centering
    \includegraphics[width=1\linewidth]{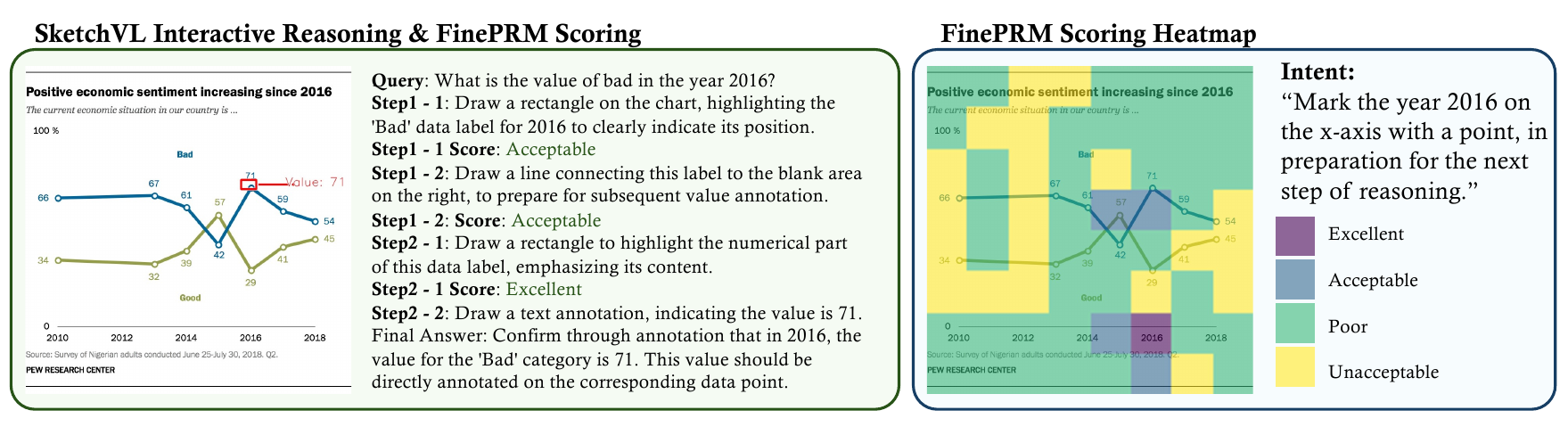}
    \caption{Left: A successful case of SketchVL's interactive reasoning, with each step scored by the FinePRM. Right: The corresponding scoring heatmap generated by the FinePRM for a specific intent.}
    \label{fig:case}
\end{figure*}

\begin{figure}
    \centering
    \includegraphics[width=1\linewidth]{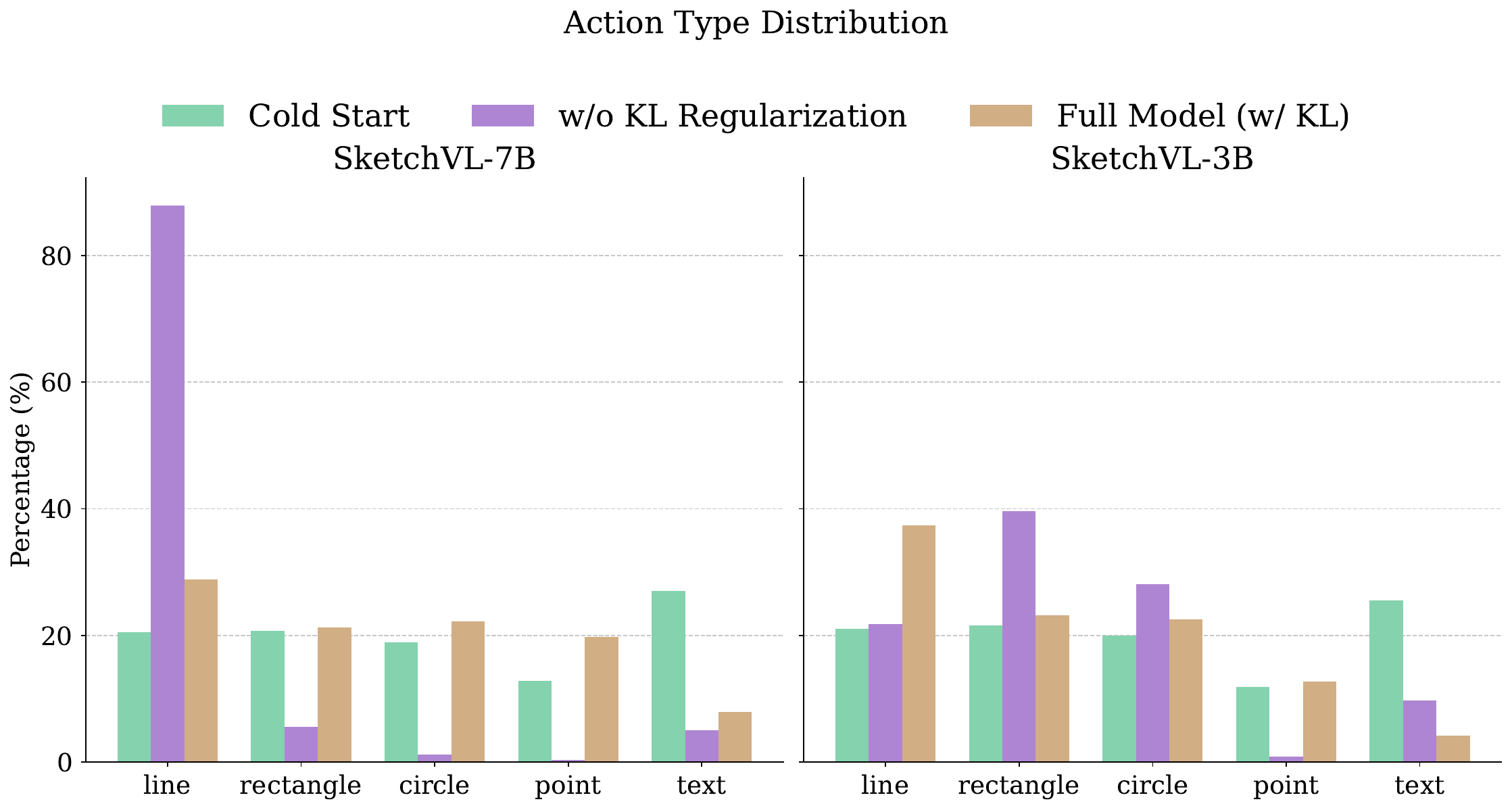}
    \caption{Effect of KL regularization on the action type distribution for SketchVL-7B (left) and SketchVL-3B (right).}

    \label{fig:hack}

\end{figure}

\subsection{Ablation Study}
\label{subsec:ablation}
To investigate the individual contributions of FinePO's core components, we conducted a series of ablation experiments using the smaller SketchVL-3B model. This allows for efficient verification of our methodological design choices.

\textbf{Value of Fine-Grained Credit Assignment in FinePO.}
The core of our experiments lies in validating the proposed fine-grained credit assignment mechanism. Compared to the baseline using naive GRPO (`w/o FinePO (naive GRPO)'), our full model achieves consistent performance gains across several key benchmarks, such as a 2.08-point improvement on ChartQA. This advantage is further substantiated by the performance drop in the `w/o FinePRM (random scores)' experiment, which confirms that our trained FinePRM provides a meaningful and necessary evaluation signal, rather than random noise. Together, these points demonstrate the value of FinePO.

\textbf{Role of KL Action Regularization.}
The `w/o KL action Regularization' model achieves performance very close to our full model. This is an expected outcome, as the KL constraint introduces a trade-off between action precision and diversity. It effectively prevents the model from collapsing to a few easy-to-score action types during training, but it can sometimes slightly constrain the optimal policy on benchmarks where a specific action type is dominant. We provide detailed analysis of this trade-off in Section~\ref{subsec:action_analysis}.

\textbf{Impact of RL and Generalization.}
As a whole, the RL phase provides a substantial performance boost. Compared to the model after only the cold-start phase, our full model shows a significant leap in performance. Notably, this improvement also extends to datasets not covered in the RL training prompts, such as PlotQA. This indicates that FinePO can enhance generalization capabilities.

\textbf{Importance of the RoI Paradigm.}
The results from the `w/o Sketch (zero GRPO)' experiment highlight the foundational role of the RoI paradigm. Removing the model's ability to draw on the image leads to a drastic collapse in performance across all benchmarks. This confirms that the iterative, visual reasoning process is fundamental for solving these complex tasks.

\subsection{Analysis of Reasoning Process}
\label{subsec:action_analysis}
To evaluate the model's quality beyond final task accuracy, we further analyze its intermediate reasoning process using our FinePRM as an automatic evaluator.

As presented in Table~\ref{tab:prm_scores}, our full models consistently achieve the highest average process scores across all test sets. This result clearly demonstrates that FinePO's fine-grained signal effectively aligns the model with the PRM. Notably, this superior process quality is demonstrated on datasets such as PlotQA, which were not included in the RL training phase, proving that FinePO enhances the model's intrinsic and generalizable reasoning capabilities.

The importance of KL action regularization is visually demonstrated in Figure~\ref{fig:hack}. While the model trained without the KL constraint improves upon the cold start model in process scores, it exhibits a severe ``Action Bias", heavily favoring a few action types. This action bias is particularly severe in the 7B model, whose action distribution almost completely collapses, likely due to its larger model capacity providing more opportunities to discover and exploit potential shortcuts or loopholes within the FinePRM. Therefore, the KL constraint does not introduce a performance trade-off in this case; instead, it prevents the model from adopting a simplistic, biased policy. By encouraging a more diverse and balanced use of actions, it compels the model to develop a more robust reasoning strategy, which ultimately leads to higher-quality intermediate steps.

\begin{table}[ht]
\centering
\caption{Average FinePRM scores on test sets for models in the ablation study. Higher scores indicate better alignment between generated actions and their intents. *\textit{Random} means that, during training, random rewards are used in place of the FinePRM.}
\small

\label{tab:prm_scores}
\begin{tabular}{@{}lccc@{}}
\toprule
Model & PlotQA & ChartBench & MMStar \\
\midrule
SketchVL-3B (Full) & \textbf{2.857} & \textbf{2.917} & \textbf{2.914} \\
 w/o FinePO (use GRPO) & 2.705 & 2.777 & 2.755 \\
 w/o FinePRM (random*) & 2.631 & 2.618 & 2.778 \\
 w/o KL Regularization & \underline{2.825} & 2.818 & 2.810 \\
 w/o RL (only cold start) & 2.747 & \underline{2.836} &\underline{2.879} \\
 \midrule
SketchVL-7B (Full) & \textbf{3.003} & \textbf{3.073} & \textbf{2.976} \\
 w/o KL Regularization & \underline{2.908} & \underline{2.921} & \underline{2.899} \\
 w/o RL (only cold start) & 2.828 & 2.897 &2.810 \\

\bottomrule
\end{tabular}
\end{table}

\subsection{Case Visualization}
\label{subsec:visualization}
Figure~\ref{fig:case} provides a qualitative visualization of SketchVL's reasoning process. The left panel illustrates a successful case where each of SketchVL's reasoning steps is scored by our FinePRM. Due to the absence of a standardized benchmark for evaluating process reward models, we assess the FinePRM's scoring capability intuitively through visualization. To produce the scoring heatmap, we define an intent, apply the action across an 8×8 image grid, and have FinePRM score each action.
The resulting heatmap demonstrates that our FinePRM can clearly distinguish between correct and incorrect regions, assigning the highest scores to the areas that most accurately fulfill the intent.

\section{Conclusion and Limitations}
In this work, we present SketchVL, a multimodal reasoning model trained with our RL algorithm, FinePO. By leveraging a FinePRM, FinePO achieves step-level credit assignment, alleviating the limitations of coarse reward signals in chart reasoning tasks. Our experiments demonstrate SketchVL's strong performance across diverse chart understanding and multimodal benchmarks. We also acknowledge the limitations of our work, which point to directions for future research. These include the current absence of a benchmark for quantitatively evaluating the FinePRM and the inherent challenge of constructing a unbiased PRM.

{
    \small
    \bibliographystyle{ieeenat_fullname}
    \bibliography{main}
}
\newpage
\section{Visualization Cases of FinePRM}
\label{sec:fineprm_vis}
\begin{figure*}[h]
    \centering
    \includegraphics[width=0.9\linewidth]{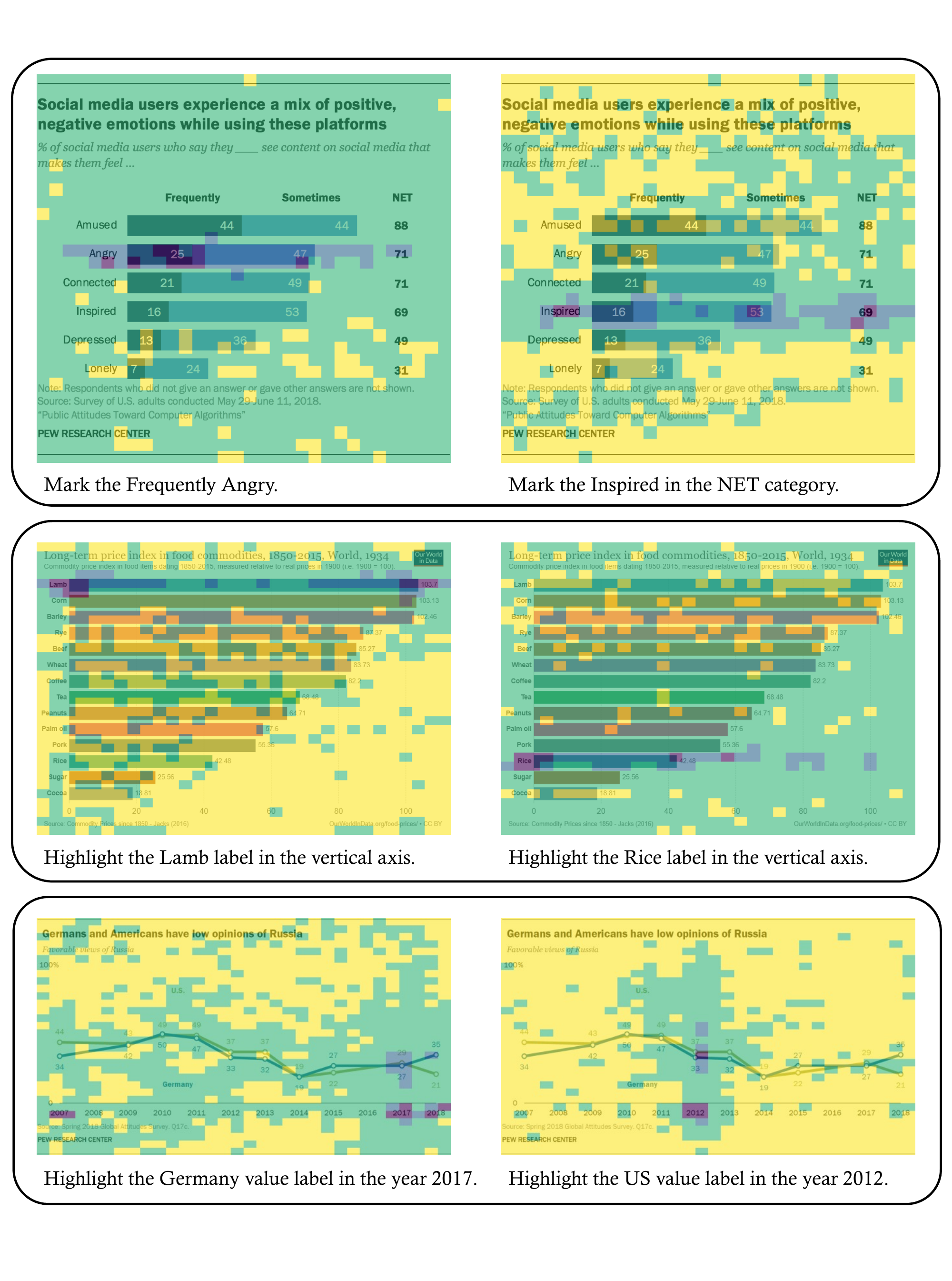}
    \caption{\textbf{Visualization of FinePRM scoring heatmaps (Case 1).} 
    We visualize the spatial credit assignment by FinePRM. The heatmap represents the score distribution across the image for a given intent. High-scoring regions indicate that FinePRM correctly identifies the areas most relevant to the instruction.}
    \label{fig:vis_case1}
\end{figure*}

\begin{figure*}[h]
    \centering
    \includegraphics[width=0.9\linewidth]{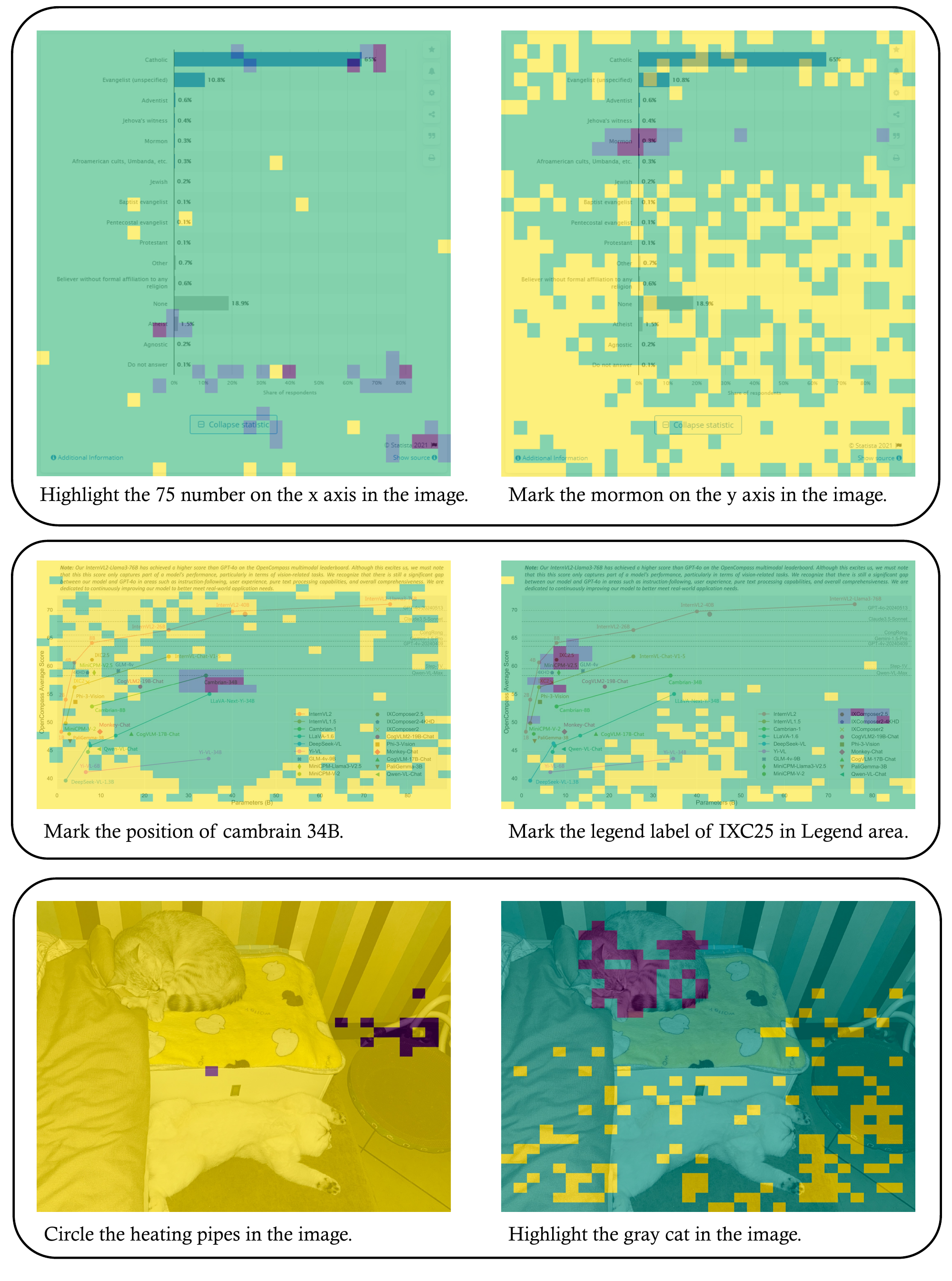}
    \caption{\textbf{Visualization of FinePRM scoring heatmaps (Case 2).} }
    \label{fig:vis_case2}
\end{figure*}

\begin{figure*}[h]
    \centering
    \includegraphics[width=0.9\linewidth]{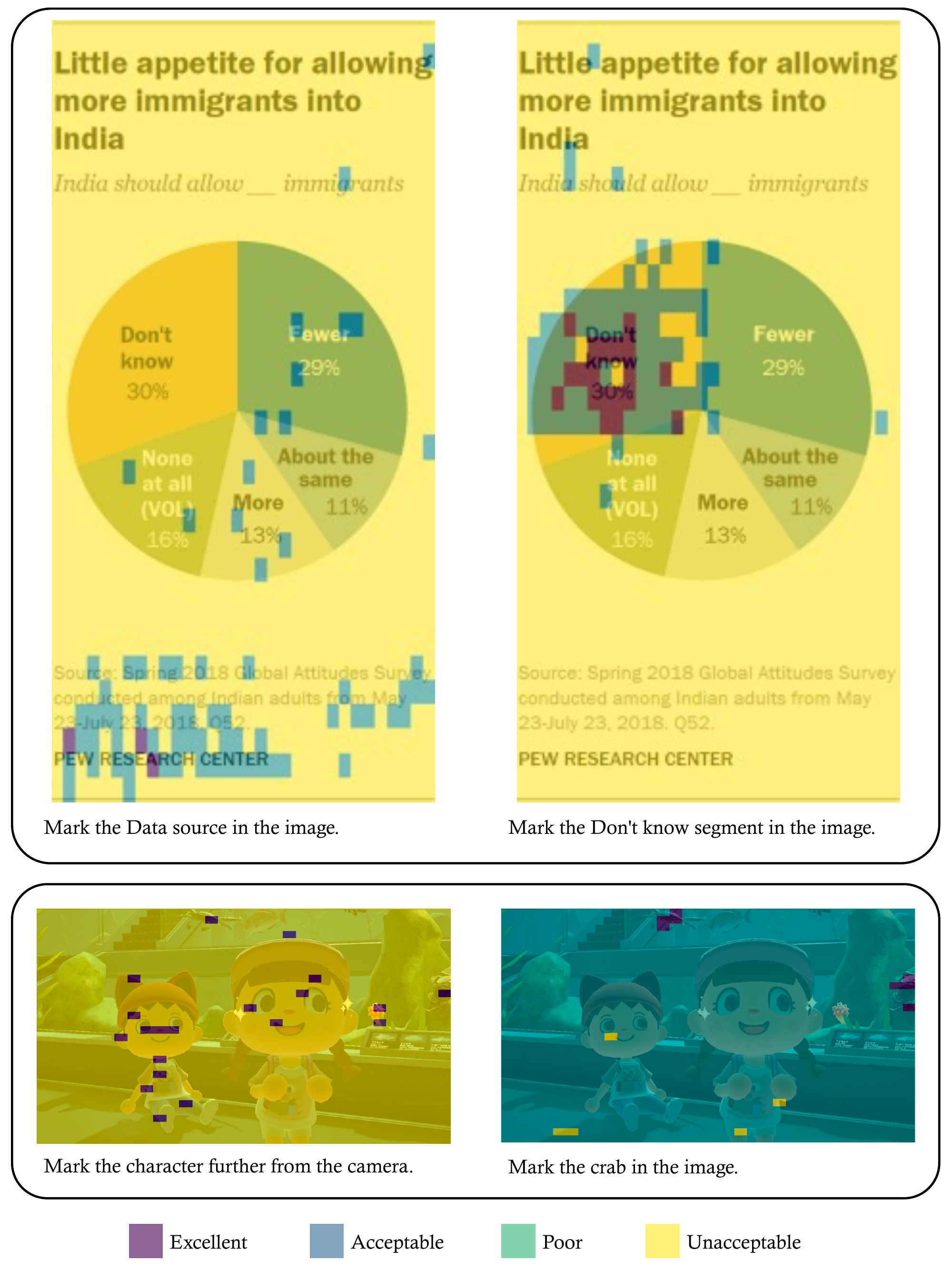}
    \caption{\textbf{Visualization of FinePRM scoring heatmaps (Case 3).} 
    Further visualization of FinePRM's scoring process on complex charts and general images. The legend indicates the score quality, where darker colors (e.g., Purple for `Excellent') represent higher scores, and lighter colors (e.g., Yellow for `Unacceptable') represent lower scores. As shown, the high-scoring regions (darker patches) align precisely with the semantic intent described below each image, confirming that FinePRM provides accurate, fine-grained feedback.}
    \label{fig:vis_case3}
\end{figure*}

In this section, we provide qualitative visualizations to assess the scoring capability of our FinePRM. As shown in Figure \ref{fig:vis_case1}, \ref{fig:vis_case2} and \ref{fig:vis_case3}, we intuitively evaluate FinePRM through visualization on complex chart and image scenarios. 

Specifically, to produce the scoring heatmaps shown in the following figures, we define a specific intent and apply the action across a $32 \times 32$ image grid. FinePRM then scores each action independently. The resulting heatmaps demonstrate that our FinePRM can clearly distinguish between correct and incorrect regions, assigning the highest scores to the areas that most accurately fulfill the intent.

\section{Prompts for Data Generation}
\label{sec:prompts}

In this section, we provide the detailed prompts used in our pipeline. To ensure the reproducibility of our method, we present the translated English versions of the prompts used for Data Enrichment, Intent-Action Synthesis (FinePRM), and Cold-Start Data Generation (SketchVL Cold Start).

\begin{tcolorbox}[colback=boxblue, colframe=frameblue, title=\textbf{Prompts for Data Enrichment and Cleaning}, fonttitle=\bfseries, boxrule=0.5mm, breakable]
\scriptsize
\textbf{1. Chart Attribute Enrichment Prompt} \\
You are a top-tier chart data annotation expert. Your task is to supplement detailed structured information for each annotation based on the given chart image and OCR-recognized text labels.
\\
\textbf{Input Format:} Normalized bounding boxes (0-1000). \\
\textbf{Task:} For each JSON object in the "Pending Annotations":
\begin{itemize}
    \item \texttt{legend\_label}: If the text relates to a legend, fill in the full legend text; otherwise, return "".
    \item \texttt{color}: Describe the color of the text or associated element (e.g., "Red", "Dark Blue").
    \item \texttt{describe}: Detailed description of the role and context (e.g., "X-axis label", "Value at the top of a bar", "Pie sector label").
\end{itemize}
\textbf{Output:} A strictly valid JSON array.

\vspace{4pt}
\hrule
\vspace{4pt}

\textbf{2. General Image Object Verification (Validity)} \\
Please refer to the two images provided (original and cropped crop with red box). Determine whether the content within the red box constitutes a meaningful, relatively complete object or a recognizable part of an object (e.g., a wheel is meaningful; a patch of solid color or sky is not).
Answer only "Yes" or "No".

\vspace{4pt}
\hrule
\vspace{4pt}

\textbf{3. General Image Fine-grained Labeling} \\
Your task is to generate a structured description for the object inside the red box to uniquely identify it in the original image.
Return JSON format:
\begin{itemize}
    \item \texttt{label}: Precise name of the object (e.g., "White porcelain plate").
    \item \texttt{relation}: Detailed description of the spatial relationship with surrounding objects to eliminate ambiguity (e.g., "In the middle of the table, below the laptop").
\end{itemize}

\vspace{4pt}
\hrule
\vspace{4pt}

\textbf{4. Duplicate Resolution Prompt} \\
\textbf{Task:} Analyze two highly overlapping annotations to determine if they point to the same object and decide which description is better. \\
\textbf{Steps:} Check the cropped images and the original image. If they are the same entity, compare "Label" and "Relation" to select the one that is more accurate and informative.
\textbf{Output:} Only the \texttt{Label} name of the retained annotation.
\end{tcolorbox}

\begin{tcolorbox}[colback=boxgreen, colframe=framegreen, title=\textbf{Prompts for Intent-Action Pair Synthesis}, fonttitle=\bfseries, boxrule=0.5mm, breakable]
\scriptsize
\textbf{System Prompt:} \\
Your task is to generate training data for a Reward Model. Based on the given image/chart configuration and annotation information, generate a JSON list containing $N$ independent "Action Pairs".

Each item must contain:
\begin{itemize}
    \item \texttt{explanation}: A natural language description of the drawing intent. It should explain \textit{what} to mark and \textit{where} it is, without revealing specific coordinates or the fact that you are looking at metadata. (e.g., "Circle the window with grilles located in the center right of the image").
    \item \texttt{action}: A single drawing instruction JSON object (`point`, `line`, `circle`, `rectangle`, or `text`) derived accurately from the provided annotations.
\end{itemize}

\textbf{Constraints:}
\begin{itemize}
    \item \textbf{Diversity:} The actions must strictly follow a specified sequence (e.g., point - line - circle) and cover different objects.
    \item \textbf{Accuracy:} Do not guess coordinates. Use the provided annotation data strictly.
    \item \textbf{Chart Specifics:} For charts, X-axis labels should be marked at the top edge center; Y-axis labels at the right edge center. Do not mark legends if they are not in the annotation.
    \item \textbf{Image Specifics:} For general images, ensure the description uniquely locates the object using spatial relationships provided in the context.
\end{itemize}

\textbf{Output Format:} A strict JSON list of objects containing `explanation` and `action` keys.
\end{tcolorbox}

\begin{tcolorbox}[colback=boxorange, colframe=frameorange, title=\textbf{Prompts for Cold-Start Data Generation}, fonttitle=\bfseries, boxrule=0.5mm, breakable]
\scriptsize
\textbf{1. Visual Question Generation} \\
Please generate a visual question regarding the chart/image content based on the provided annotations. The question should:
\begin{itemize}
    \item Focus on the visual content (e.g., values, trends, object relationships, counting).
    \item Be concise and answerable based strictly on the image.
    \item (For Charts) Focus on topics like: Extremes \& Ranking, Comparisons, Trends, Statistics.
    \item (For Images) Focus on topics like: Spatial relationships, Attributes, Actions, Context.
\end{itemize}
\textbf{Output:} Only the question string.

\vspace{4pt}
\hrule
\vspace{4pt}

\textbf{2. Sketch-CoT Reasoning Generation} \\
Your task is to answer the user's question by generating a multi-step reasoning process that includes visual sketching actions.

\textbf{Output Format:} A JSON list of lists, where each sub-list represents a reasoning turn.
\begin{itemize}
    \item \texttt{explanation}: Summarize the previous step or plan the current drawing. Do not state the final answer until the last step.
    \item \texttt{action}: A dictionary defining a drawing operation (`point`, `line`, `circle`, `rectangle`, `text`) with normalized coordinates (0-1000).
\end{itemize}

\textbf{Key Instructions:}
\begin{itemize}
    \item \textbf{Draw then Conclude:} You should first perform a drawing action (visual grounding) and then discuss the finding in the next turn.
    \item \textbf{No Hallucination:} Use strictly existing coordinates from the provided metadata.
    \item \textbf{Chain of Thought:} The reasoning should be linear. The first turn plans the path; intermediate turns verify data points via drawing; the final turn (with empty action) provides the direct answer to the question.
    \item \textbf{Conciseness:} Solve the problem in typically 3-4 turns.
\end{itemize}

\textbf{Input Provided:} Chart HTML/Image Annotations and the User Query.
\end{tcolorbox}

\onecolumn 

\section{Implementation of Fine-Grained Credit Assignment}
\label{sec:algorithm_impl}

In this section, we provide the implementation logic for our proposed Fine-Grained Policy Optimization (FinePO). Due to the complexity of the logic, we present the code in a single-column format.

\subsection{Variable Definitions}
To facilitate the understanding of the provided Python implementation, we map the key variables in the code to the mathematical notations defined in the FinePO methodology.

\begin{itemize}
    \item \texttt{inputs}: The batch data structure containing the generated trajectories $\{y_1, \dots, y_k\}$, their corresponding FinePRM process scores $p_j$, and associated metadata.
    
    \item \texttt{prior\_action\_distribution}: Represents the fixed prior distribution $Q(a)$ derived from the training set, serving as the anchor for the KL divergence constraint.
    
    \item \texttt{action\_history}: A sliding window buffer used to estimate the current policy's action distribution $P_k(a)$, ensuring a stable calculation for the dynamic KL penalty.
    
    \item \texttt{kl\_lambda}: Corresponds to the coefficient $\lambda_{\text{KL}}$. It controls the strength of the penalty applied when the current action distribution deviates from the prior.
    
    \item \texttt{reward\_offsets}: Stores the computed clipped KL offset values $\mathcal{O}_{\text{clipped}}(a_j)$. These offsets are added to the raw FinePRM scores to regularize the action types.
    
    \item \texttt{scalar\_adv}: Represents the coarse, cross-trajectory advantage $A(y_i)$. This scalar serves as the base value and the total "budget" to be redistributed among the individual steps.
    
    \item \texttt{s\_avg}: The length-weighted average score $\bar{p}$ of all steps within a single response. It acts as the intra-response baseline to determine the relative quality $\Delta_j$ of each step.
    
    \item \texttt{k}: The dynamic scaling factor $k$. It scales the fine-grained adjustment based on the magnitude of the coarse advantage $|A(y_i)|$ and the maximum score deviation.
    
    \item \texttt{alpha}: Corresponds to the hyperparameter $\alpha$. It governs the intensity of the redistribution, determining how much the fine-grained scores influence the final token-level advantage.
    
    \item \texttt{token\_spans}: A utility mapping used to determine the token length $L_j$ and specific indices for each reasoning step $s_j$, enabling the precise application of the final step-level advantage $A(s_j)$ to the token sequence.
\end{itemize}

\begin{tcolorbox}[colback=white, colframe=gray!50!black, title=\textbf{Algorithm 1: Fine-Grained Advantage Redistribution}, fonttitle=\bfseries, boxrule=0.5mm, breakable]
\scriptsize
\begin{verbatim}
def _prepare_batch_inputs(self, inputs: DataType) -> List[DataType]:
    """
    Prepare batch inputs with KL penalty and Token-Level Credit Assignment.
    This function runs on each worker process.
    """
    # --- Part 1: Global Distribution Alignment (KL Penalty) ---
    credit_key = f'credit_details_{self.reward_func_names[0]}'
    
    # 1. Gather local actions from the current batch
    local_actions = []
    for data in inputs:
        if credit_key in data:
            # Extract action types (e.g., 'point', 'line') from step strings
            for step_str in data[credit_key].keys():
                action = self._parse_action_type(step_str)
                if action in self.prior_action_distribution:
                    local_actions.append(action)

    # 2. Synchronize action counts across all distributed workers
    # gathered_actions_flat_list contains actions from ALL GPUs
    global_actions = gather_object(local_actions)
    reward_offsets = {}

    # 3. Calculate KL-based Reward Offsets (Executed on Main Process)
    if self.accelerator.is_main_process:
        # Update sliding window history with current global counts
        global_counts = collections.Counter(global_actions)
        self.action_history.append(global_counts) 
        
        # Aggregate counts from the sliding window
        total_counts = collections.defaultdict(int)
        for counts in self.action_history:
            for act, c in counts.items():
                total_counts[act] += c
        total_n = sum(total_counts.values())
        
        if total_n > 0:
            # Calculate current distribution P_curr
            p_current = {a: c / total_n for a, c in total_counts.items()}
            
            # Calculate Offset: -lambda * log(P_curr / P_prior)
            for action, p_prior in self.prior_action_distribution.items():
                p_curr = p_current.get(action, 0.0) + self.epsilon
                # KL Divergence direction: Penalty if P_curr > P_prior
                offset = -self.kl_lambda * math.log(p_curr / (p_prior + self.epsilon))
                # Clip offset to prevent training instability
                reward_offsets[action] = clip(offset, -self.clip_val, self.clip_val)

    # 4. Broadcast calculated offsets to all workers
    broadcast_object_list([reward_offsets], from_process=0)
    
    # 5. Apply offsets to local data (Update Reward Scores)
    for data in inputs:
        if credit_key in data:
            for step_str, score in data[credit_key].items():
                action = self._parse_action_type(step_str)
                # Add the global penalty/bonus to the step score
                data[credit_key][step_str] = score + reward_offsets.get(action, 0.0)

    # --- Part 2: Token-Level Credit Assignment ---
    # Split data into mini-batches for memory efficiency
    batches = self.split_by_mini_batches(inputs)
    processed_batches = []
    
    for batch in batches:
        # (Standard encoding and masking logic omitted for brevity...)
        # ...
        
        # Initialize a tensor to hold per-token advantages
        # Shape: [Batch_Size, Sequence_Length]
        advantages_tensor = torch.zeros_like(labels, dtype=torch.float32)
        
        for j, sample in enumerate(batch):
            scalar_adv = sample['advantages'].item() # Original GAE advantage
            credit_details = sample.get(credit_key, {})
            
            # Identify valid completion tokens
            mask = (labels[j] != -100)
            
            # Default: Assign the scalar advantage to ALL tokens in the response
            per_token_adv = torch.full((mask.sum(),), scalar_adv)
            
            # If FinePRM scores exist, perform redistribution
            if credit_details:
                # 1. Map logical steps to physical token indices
                # Returns: {'step_1': (start_idx, end_idx), ...}
                token_spans = self._map_sub_steps_to_tokens(sample, credit_details)
                
                # 2. Calculate Length-Weighted Average Score (S_avg)
                w_score_sum = 0
                total_len = 0
                for step_str, score in credit_details.items():
                    if step_str in token_spans:
                        span_len = token_spans[step_str].len
                        w_score_sum += score * span_len
                        total_len += span_len
                
                s_avg = w_score_sum / total_len if total_len > 0 else 0
                
                # 3. Determine Dynamic Scaling Factor (k)
                # We calculate deviations from the average
                offsets = [score - s_avg for score in credit_details.values()]
                max_deviation = max(offsets) if offsets else 0
                
                if max_deviation > 0:
                    # k scales the deviation so that the max deviation roughly 
                    # corresponds to the magnitude of the scalar advantage.
                    k = abs(scalar_adv) / (max_deviation + 1e-6)
                else:
                    k = 0

                # 4. Redistribute Advantage
                for step_str, score in credit_details.items():
                    if step_str not in token_spans: continue
                    
                    start, end = token_spans[step_str]
                    offset = score - s_avg
                    
                    # Core Redistribution Formula: 
                    # A_new = A_original + alpha * k * (Score_step - Score_avg)
                    adjustment = self.alpha * k * offset
                    mod_adv = scalar_adv + adjustment
                    
                    # 5. Conservative Clipping (Advantage Conservation)
                    # Ensure the sign doesn't flip excessively and stays within bounds
                    if scalar_adv < 0: # Negative feedback case
                        # Allow worse steps to be more negative, but limit improvement
                        lower_bound = 2.0 * scalar_adv
                        mod_adv = max(lower_bound, min(mod_adv, 0.0))
                    else: # Positive feedback case
                        # Allow better steps to be more positive, but limit punishment
                        upper_bound = 2.0 * scalar_adv
                        mod_adv = max(0.0, min(mod_adv, upper_bound))
                        
                    # Assign modified advantage to the specific token span
                    per_token_adv[start:end] = mod_adv

            # Fill the tensor
            advantages_tensor[j, mask] = per_token_adv

        # Update the batch data structure
        encoded_batch['advantages'] = advantages_tensor
        processed_batches.append(encoded_batch)
        
    return processed_batches
\end{verbatim}
\end{tcolorbox}

\twocolumn
\section{Evaluation Prompts}
\label{sec:eval_prompts}

To ensure a fair and rigorous comparison, we employ a model-based judge (DeepSeek-R1-Distill-Qwen-14B) to evaluate the correctness of SketchVL's responses. Depending on the dataset, we apply specific evaluation criteria to handle the nuances of chart reasoning. Below are the translated system prompts used for each benchmark.

\begin{tcolorbox}[colback=boxpurple, colframe=framepurple, title=\textbf{Prompt for ChartQA-Pro Evaluation}, fonttitle=\bfseries, boxrule=0.5mm]
\scriptsize
\textbf{System Instruction:} \\
Please judge the correctness of the following answer. Briefly analyze or calculate first, and finally append a \texttt{[true]} tag if correct, or \texttt{[false]} if incorrect. 
Focus solely on whether the final conclusion matches the ground truth label. Do not judge the reasoning process.

\textbf{Evaluation Principles (Relaxed Correctness Metric):}
\begin{enumerate}
    \item \textbf{MCQ \& Fact Checking:} Use \textbf{Exact Match}. The predicted option (A/B/C/D) or Boolean (True/False) must strictly match the label.
    \item \textbf{Year Values:} Use \textbf{Exact Match}. Years (e.g., 2010 vs 2009) must be precise; visual approximation errors are not allowed for years.
    \item \textbf{Other Numeric Values:} Allow a \textbf{5\% relative error}. Calculate error $E = |Pred - GT| / |GT|$. If $E \leq 0.05$, mark as correct. Handle unit conversions if necessary.
    \item \textbf{Textual Answers:} Assess \textbf{Semantic Similarity}. Exact match is not required; ignore minor differences like pluralization ('Female' vs 'Females') or casing.
    \item \textbf{List Answers:} Split the list and evaluate each element individually using rules 1-4. All elements must be correct.
\end{enumerate}
\end{tcolorbox}

\begin{tcolorbox}[colback=boxpurple, colframe=framepurple, title=\textbf{Prompt for PlotQA Evaluation}, fonttitle=\bfseries, boxrule=0.5mm]
\scriptsize
\textbf{System Instruction:} \\
Analyze the correctness of the response. Output \texttt{[true]} or \texttt{[false]} at the end. Do not provide an ambiguous "neither" option.

\textbf{Evaluation Principles:}
\begin{itemize}
    \item \textbf{Non-Numeric Answers:} Analyze semantic understanding. Colors are correct if they are semantically similar.
    \item \textbf{Numeric Answers:} You must perform a calculation to verify the error. Compute accuracy $acc = |Label - Answer| / |Label|$.
    \item \textbf{Threshold:} If $acc \leq 0.05$, evaluate as True; otherwise False.
    \item \textbf{Units:} If units do not align, convert them based on context before calculating the error.
    \item \textbf{Formulas:} If the assistant outputs an uncomputed formula, evaluate its final result against the label with the 5\% tolerance.
\end{itemize}
\end{tcolorbox}

\begin{tcolorbox}[colback=boxpurple, colframe=framepurple, title=\textbf{Prompt for EvoChart Evaluation}, fonttitle=\bfseries, boxrule=0.5mm]
\scriptsize
\textbf{System Instruction:} \\
Judge the correctness based on the provided content. Output \texttt{[true]} or \texttt{[false]}.

\textbf{Evaluation Principles:}
\begin{itemize}
    \item \textbf{Condition \texttt{is\_clear}:} If \texttt{is\_clear=true} is specified in metadata, require strictly precise numerical correspondence.
    \item \textbf{Condition \texttt{not is\_clear}:} If \texttt{is\_clear=false}, allow a 5\% tolerance for all numeric labels.
    \item \textbf{Calculation:} For numeric labels (e.g., 27 vs 28), if the relative error is within $\pm 5\%$, count it as correct. Example: $(22-21)/21 < 0.05$ is acceptable.
    \item \textbf{API Errors:} If an API error occurs in the model output, look for a correct answer generated prior to the error.
\end{itemize}
\end{tcolorbox}

\begin{tcolorbox}[colback=boxpurple, colframe=framepurple, title=\textbf{Prompt for General, Math and Other Evaluation}, fonttitle=\bfseries, boxrule=0.5mm]
\scriptsize
\textbf{System Instruction:} \\
Judge the correctness of this subjective or general question. Output \texttt{[true]} or \texttt{[false]}.

\textbf{Evaluation Principles:}
\begin{itemize}
    \item \textbf{Numeric Answers:} Allow a 5\% tolerance excluding units.
    \item \textbf{Years:} Be lenient. Treat years as numeric values with 5\% tolerance (e.g., 2018 vs 2022 is within 5\% relative error).
    \item \textbf{Non-Numeric:} Colors and text need only be semantically close.
    \item \textbf{Objective:} Focus on the final answer, ignoring intermediate reasoning steps unless they contradict the correct final result.
\end{itemize}
\end{tcolorbox}

\section{Training Dynamics Analysis}
\label{sec:training_dynamics}

\begin{figure*}[t]
    \centering
    \begin{subfigure}{0.49\linewidth}
        \centering
        \includegraphics[width=\linewidth]{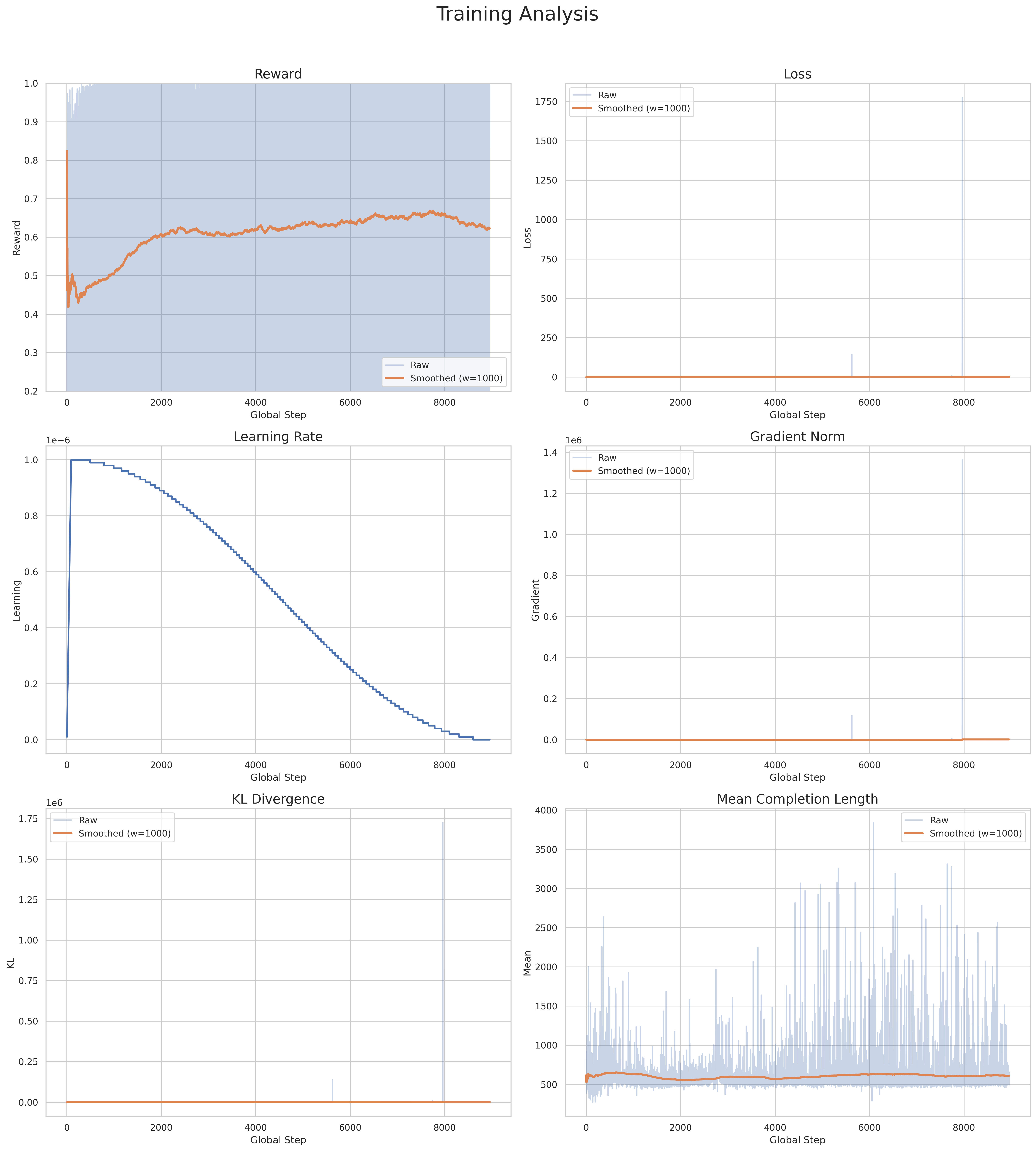}
        \caption{\textbf{Full SketchVL-3B (Ours).} Trained with FinePO and FinePRM.}
        \label{fig:curve_full}
    \end{subfigure}
    \hfill
    \begin{subfigure}{0.49\linewidth}
        \centering
        \includegraphics[width=\linewidth]{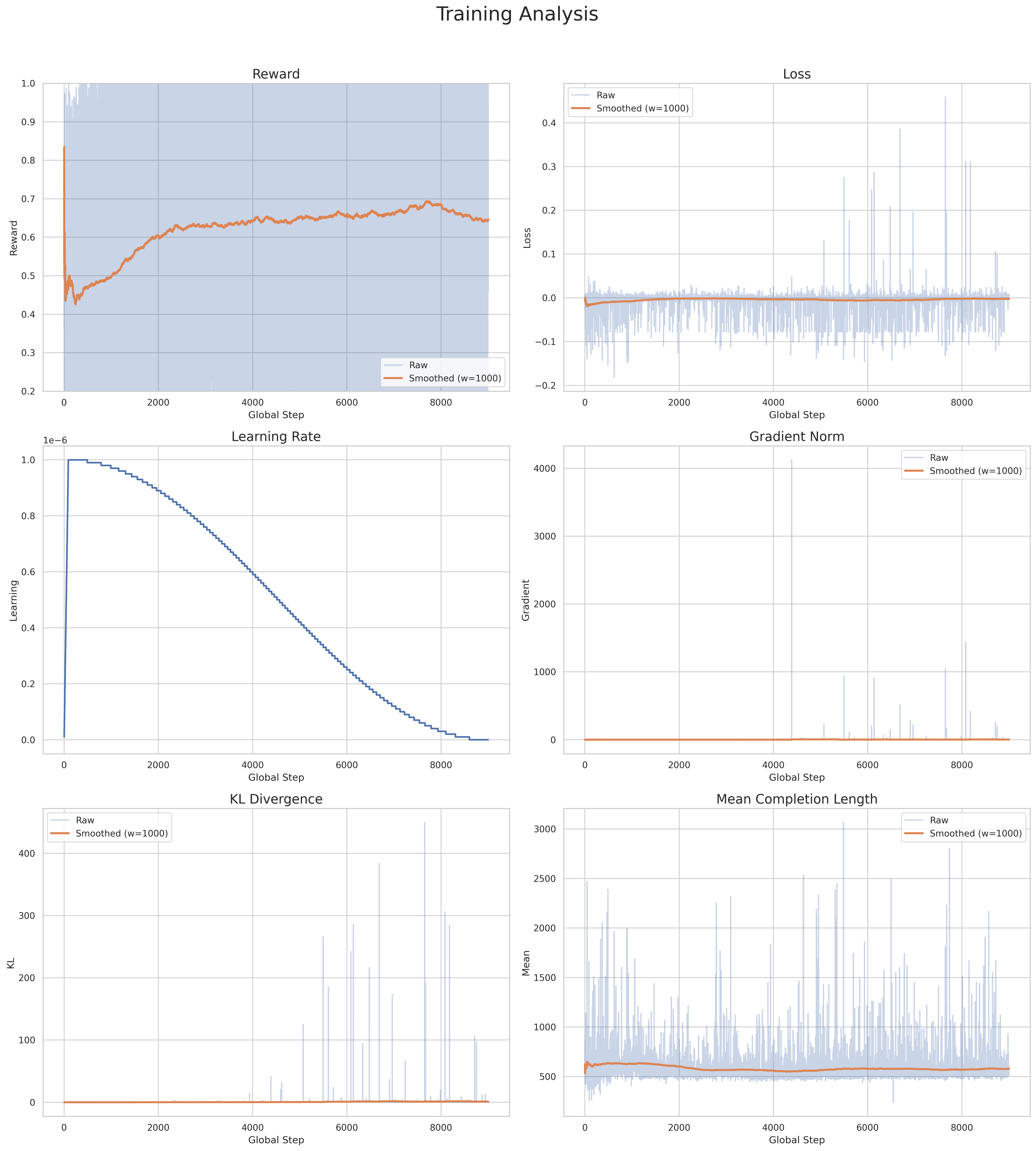}
        \caption{\textbf{w/o FinePO (Naive GRPO).} Uses only coarse trajectory-level rewards.}
        \label{fig:curve_nofinepo}
    \end{subfigure}
    
    \vspace{1em} 

    \begin{subfigure}{0.49\linewidth}
        \centering
        \includegraphics[width=\linewidth]{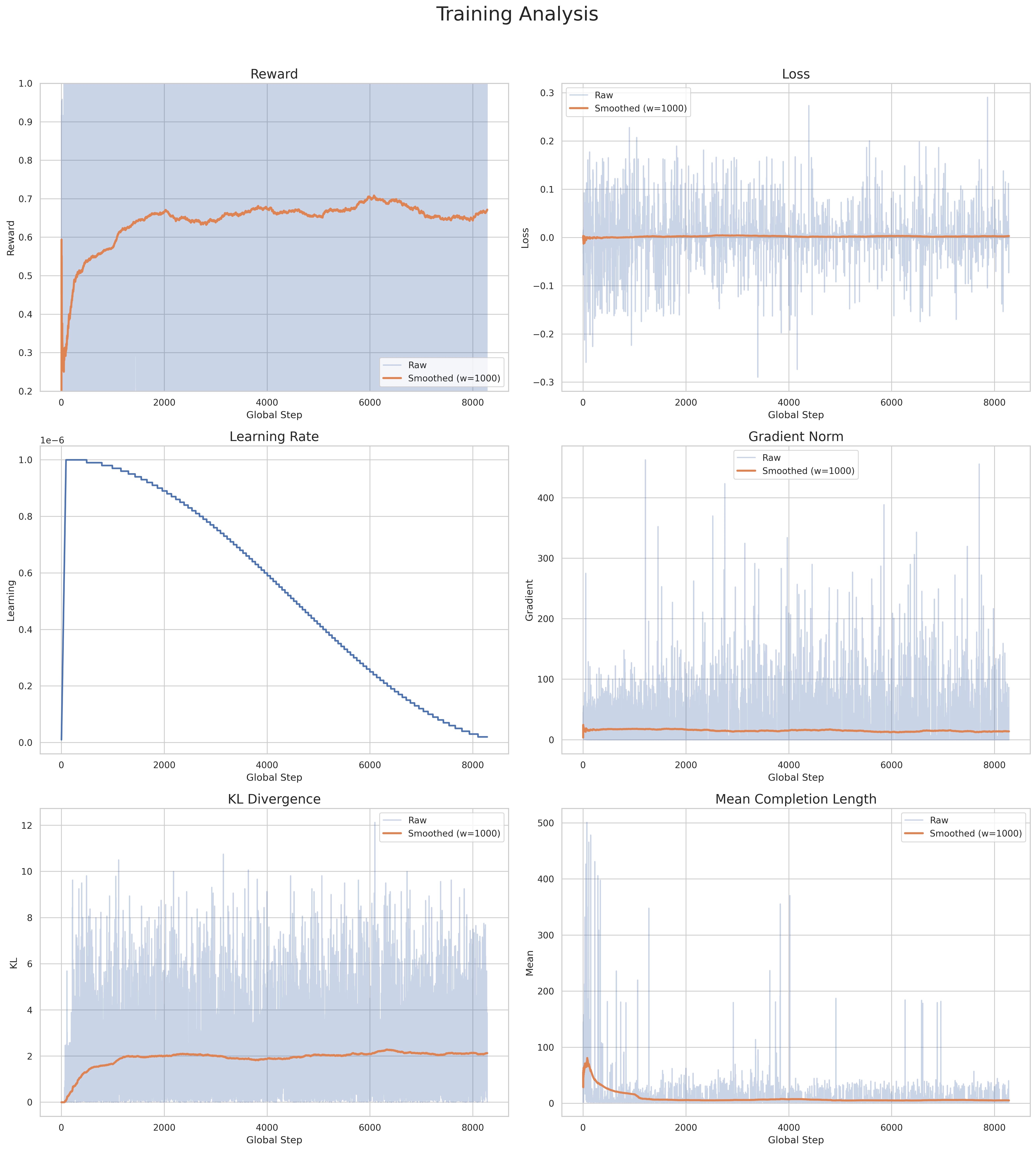}
        \caption{\textbf{w/o Sketch (Zero GRPO).} No interactive reasoning actions.}
        \label{fig:curve_nosketch}
    \end{subfigure}
    \hfill
    \begin{subfigure}{0.49\linewidth}
        \centering
        \includegraphics[width=\linewidth]{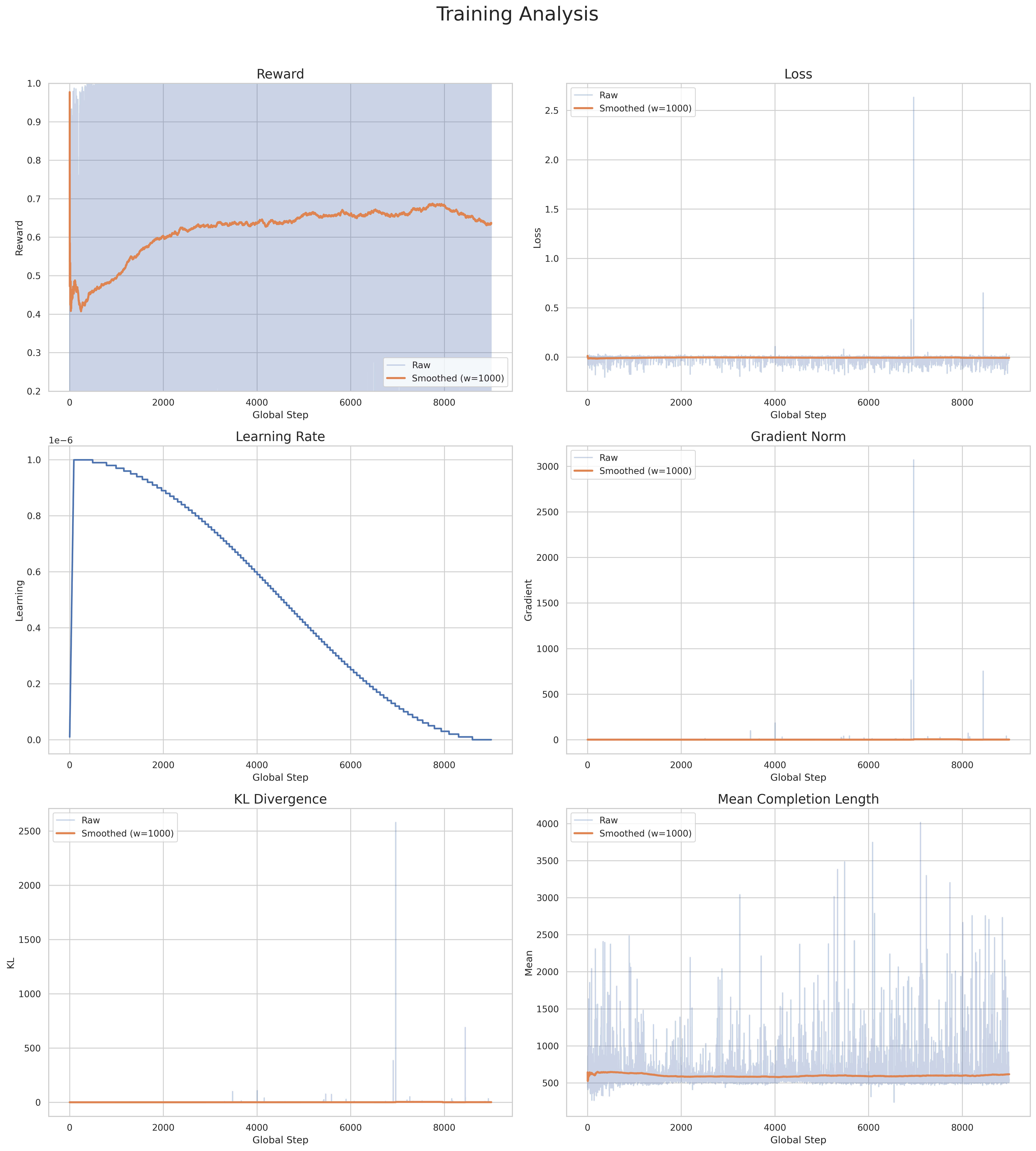}
        \caption{\textbf{w/o FinePRM (Random Scores).} Credit is assigned randomly.}
        \label{fig:curve_rand}
    \end{subfigure}
    
    \caption{\textbf{Comparison of Training Dynamics.} We visualize the training metrics for the full SketchVL-3B model and three ablation baselines. The curves include Reward (mean training reward), Loss, Learning Rate, Gradient Norm, KL Divergence (between the current policy and the reference model), and Mean Completion Length.}
    \label{fig:training_curves}
\end{figure*}

To further validate the effectiveness of our method, we visualize the training dynamics of the SketchVL-3B model and its ablation variants. \Cref{fig:training_curves} presents the curves for Reward, Loss, Learning Rate, Gradient Norm, KL Divergence, and Mean Completion Length over the training steps.

\paragraph{Stability Analysis.}
As observed in the \textbf{Gradient Norm} and \textbf{KL Divergence} curves, the full SketchVL-3B model (\Cref{fig:curve_full}) exhibits significantly higher training stability compared to the baselines. 
\begin{itemize}
    \item The \textbf{w/o FinePO (naive GRPO) } model (\Cref{fig:curve_nofinepo}), which relies on coarse trajectory-level rewards (Naive GRPO), shows drastic spikes in Gradient Norm and unstable KL divergence. This confirms that uniformly broadcasting rewards to all tokens introduces substantial noise, destabilizing the policy update.
    \item Similarly, the \textbf{w/o Sketch (zero GRPO)} model (\Cref{fig:curve_nosketch}) lacks the explicit grounding of reasoning steps, leading to a harder optimization landscape. Notably, although it achieves a relatively ideal training reward, it performs poorly on test benchmarks. 
\end{itemize}
In contrast, our FinePO mechanism (\Cref{fig:curve_full}) effectively smooths the learning process by assigning precise credit to individual steps, resulting in stable gradient updates and controlled policy deviation.

\paragraph{Reward Analysis.}
It is worth noting that the final converged Reward value of the full SketchVL model is slightly lower than that of the baselines (e.g., w/o FinePO). This is an expected and reasonable outcome.
Our FinePO algorithm introduces strict constraints—specifically the \textbf{KL Action Regularization} and the \textbf{FinePRM alignment}—to prevent "reward hacking" (where the model exploits loopholes to get high scores without proper reasoning).
While these additional rules and constraints slightly reduce the absolute scalar value of the reward, they force the model to adhere to a more rigorous and generalizable reasoning logic, ultimately leading to better performance on actual benchmarks as shown in our main experiments.


\end{document}